\documentclass{bmvc2k}
\usepackage{multirow}

\title{Improving Heterogeneous Face Recognition \\ with Conditional Adversarial Networks}

\addauthor{Wuming Zhang}{wuming.zhang@ec-lyon.fr}{1}
\addauthor{Zhixin Shu}{zhshu@cs.stonybrook.edu}{2}
\addauthor{Dimitris Samaras}{samaras@cs.stonybrook.edu}{2}
\addauthor{Liming Chen}{liming.chen@ec-lyon.fr}{1}

\addinstitution{
 Laboratory LIRIS\\
 Ecole Centrale de Lyon\\
 Ecully, France
}
\addinstitution{
 Computer Vision Lab\\
 Stony Brook University\\
 Stony Brook, NY, USA
}

\runninghead{W. ZHANG ET AL.}{Improving HFR with CGAN}

\newsavebox{\measurebox}

\begin{document}

\maketitle

\begin{abstract}
Heterogeneous face recognition between color image and depth image is a much desired capacity for real world applications where shape information is looked upon as merely involved in gallery. In this paper, we propose a cross-modal deep learning method as an effective and efficient workaround for this challenge. Specifically, we begin with learning two convolutional neural networks (CNNs) to extract 2D and 2.5D face features individually. Once trained, they can serve as pre-trained models for another two-way CNN which explores the correlated part between color and depth for heterogeneous matching. Compared with most conventional cross-modal approaches, our method additionally conducts accurate depth image reconstruction from single color image with Conditional Generative Adversarial Nets (cGAN), and further enhances the recognition performance by fusing multi-modal matching results. Through both qualitative and quantitative experiments on benchmark FRGC 2D/3D face database, we demonstrate that the proposed pipeline outperforms state-of-the-art performance on heterogeneous face recognition and ensures a drastically efficient on-line stage.
\end{abstract}

\section{Introduction}
\label{sec:intro}
For decades, face recognition (FR) from color images has achieved substantial progress and forms part of an ever-growing number of real world applications, such as video surveillance, people tagging and virtual/augmented reality systems \cite{zhao2003face, tan2006face, azeem2014survey}. With the increasing demand for recognition accuracy under unconstrained conditions, the weak points of 2D based FR methods become apparent: as an imaging-based representation, color image is quite sensitive to numerous external factors, such as lighting variations and makeup patterns. Therefore, 3D based FR techniques \cite{ding2016comprehensive, corneanu2016survey, bowyer2006survey} have recently emerged as a remedy because they take into consideration the intrinsic shape information of faces which is more robust while dealing with these nuisance factors. Moreover, the complementary strengths of color and depth data allow them to jointly work and gain further improvement. 

However, depth data is not always accessible in real-life conditions due to its special requirements for optical instruments and acquisition environment. Likewise, other challenges remain as well, including the real-time registration and preprocessing for depth images. An important question then naturally arises: can we design a recognition pipeline where depth images are only registered in gallery while still providing significant information for the identification of unseen color images? To cope with this problem, heterogeneous face recognition (HFR) \cite{toderici2010bidirectional, zhao2013benchmarking, huang2012oriented} has been proposed as a reasonable workaround. As a worthwhile trade-off between purely 2D and 3D based method, HFR adopts both color and depth data for training and gallery set while the online probe set will simply contains color images. Under this mechanism, a HFR framework can take full advantage of both color and depth information at the training stage to reveal the correlation between them. Once learned, this cross-modal correlation makes it possible to conduct heterogeneous matching between preloaded depth images in gallery and color images digitally captured in real time.

Beyond the above-mentioned mechanism, in this paper we take a further look at our constraint on the use of depth image. Note that all difficulties, which hinder us from availing ourselves of depth information in probe set, come from the acquisition and registration of 3D data. Intuitively, these problems can be immediately solved if we can reconstruct depth image from color image accurately and efficiently. Despite many existing work on shape recovery from single image, most of them rely on 3D model fitting which is time-consuming and can be prone to lack accuracy when landmarks are not precisely located. Thanks to the extremely rapid development of generative models, especially the Generative Adversarial Network (GAN) \cite{goodfellow2014generative} and its conditional variation (cGAN) \cite{mirza2014conditional} which are introduced quite recently, we implement an end-to-end depth face recovery with cGAN to enforce the realistic image generation. Furthermore, the recovered depth information enables a straightforward comparison in 2.5D space. 

\begin{figure}
\includegraphics[width=\linewidth]{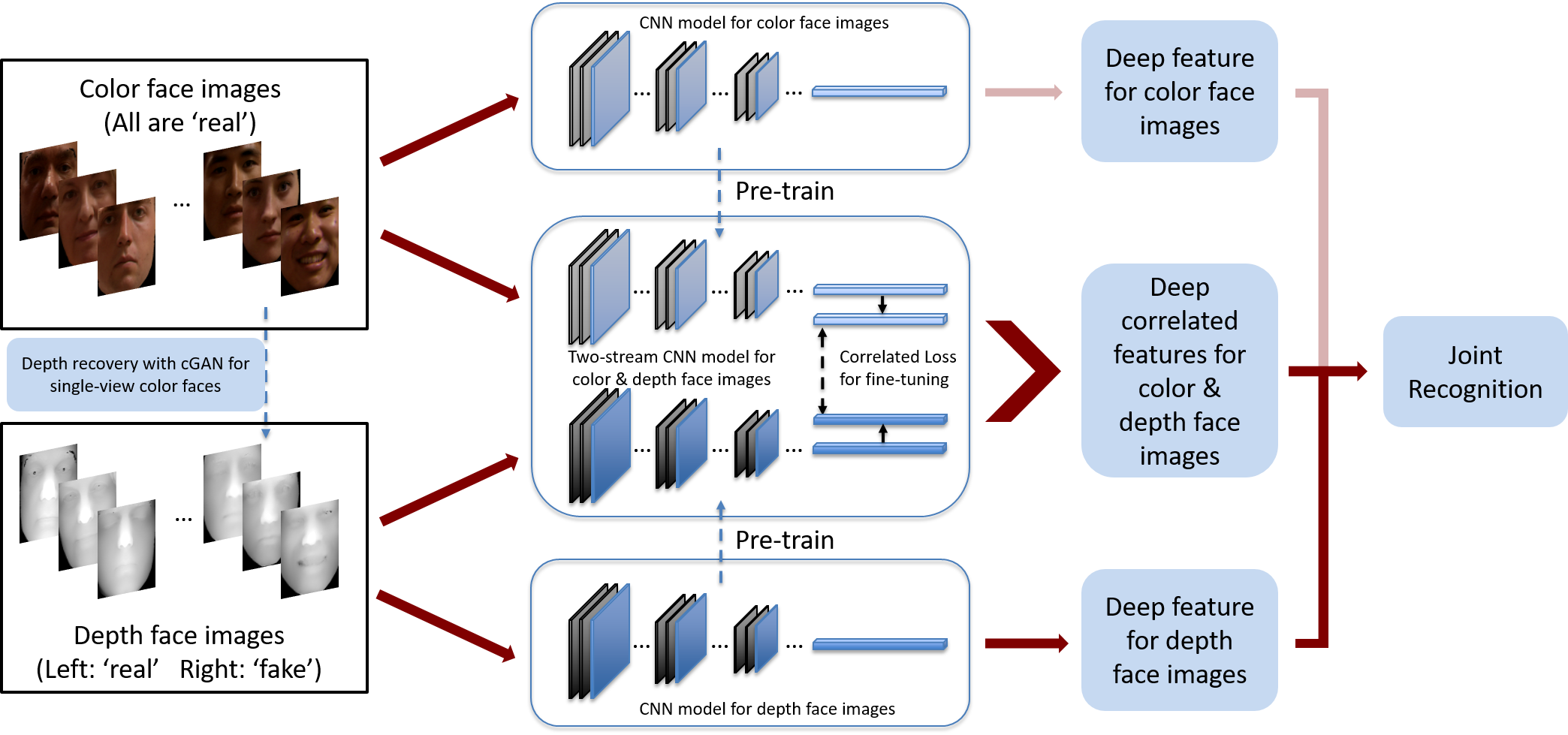}
\caption{Overview of the proposed CNN models for heterogeneous face recognition. Note that (1) depth recovery is conducted only for testing; (2) the final joint recognition may or may not include color based matching, depending on the specific experiment protocol. }
\label{fig:overview}
\end{figure}

A flowchart of the proposed method is illustrated in Fig. \ref{fig:overview}, and we list our contributions as follows:
\begin{itemize}
\item A novel depth face recovery method based on cGAN and Auto-encoder with skip connections which greatly improves the quality of reconstructed depth images.
\item We first train two discriminative CNNs individually for a two-fold purpose: to extract features of color image and depth image, and to provide pre-trained models for the cross-modal 2D/2.5D CNN model. 
\item A novel heterogeneous face recognition pipeline which fuses multi-modal matching scores to achieve state-of-the-art performance.
\end{itemize}

\section{Related Work}
\subsection{3D Face Reconstruction}

3D face reconstruction from single/multiple images or stereo video has been a challenging task due to its nonlinearity and ill-posedness. A number of prevailing approaches addressed this problem based on shape-subspace projections, where a set of 3D prototypes are fitted by adjusting corresponding parameters to a given 2D image and most of them were derived from 3DMM \cite{blanz2003face} and Active Appearance Models \cite{matthews20072d}. Alternative models were afterwards proposed as well which follow the similar processing pipeline by fitting 3D models to 2D images through various face collections or prior knowledge. For example, Gu and Kanade \cite{gu20063d} fit surface 3D points and related textures together with the pose and deformation estimation. Kemelmacher-Shlizerman et al. \cite{kemelmacher20113d} considered the input image as a guide with a single reference model to achieve 3D reconstruction. In recent work of Liu et al. \cite{liu2016joint}, two sets of cascaded regressors are implemented and correlated via a 3D-2D mapping iteratively to solve face alignment and 3D face reconstruction simultaneously. Likewise, using generic model remains a decent solution as well for 3D face reconstruction from stereo videos, as presented in \cite{chowdhury20023d, fidaleo2007model,park20073d}. Despite of strikingly accurate reconstruction result reported in the above researches, the drawback of relying on single or a large number of well-aligned 3D training data is observed and even enlarged here, because as far as we know 3D prototypes are necessary for almost all reconstruction approaches.

\subsection{2D-3D Heterogeneous Face Recognition}
As a pioneer and cornerstone for numerous subsequent 3D Morphable Model (3DMM) based methods, Blanz and Vetter \cite{blanz2003face} built this statistical model by merging a branch of 3D face models and then densely fit it to a given facial image for further matching. Toderici et al. \cite{toderici2010bidirectional} located some predefined key landmarks on the facial images in different poses, and then roughly align them to a frontal 3D model to achieve recognition target; Riccio and Dugelay \cite{riccio2007geometric} also established a dense correspondence between the 2D probe and the 3D gallery using geometric invariants across face region. Following this framework, a pose-invariant asymmetric 2D-3D FR approach \cite{zhang20123d} was proposed which conducts a 2D-2D matching by synthesizing 2D image from corresponding 3D models towards the same pose as a given probe sample. This approach was further extended and compared with work of Zhao et al. \cite{zhao2013benchmarking} as a benchmarking asymmetric 2D-3D FR system, a complete version of their work was recently released in \cite{kakadiaris20163d}. Though the above models achieved satisfactory performance, unfortunately they all suffer from high computational cost and long convergence process owing to considerable complexity of pose synthesis, and their common assumption that accurate landmark localization in facial images was fulfilled turns out to be another tough topic. More recently, learning based approaches have significantly increased on 2D/3D FR. Huang et al. \cite{huang2012oriented} projected the proposed illuminant-robust feature OGM onto the CCA space to maximize the correlation between 2D/3D features; instead, Wang et al. \cite{wang20142d} combined Restricted Boltzmann Machines (RBMs) and CCA/kCCA to achieve this goal. The work of Jin et al. \cite{jin2014cross}, called MSDA based on Extreme Learning Machine (ELM) as aforementioned, aims at finding a common discriminative feature space revealing the underlying relationship between different views. These approaches take well advantage of learning model, but would encounter weakness when dealing with non-linear manifold representations.

\section{Depth Face Reconstruction}
The target of taking a random color face image to recover its counterpart in depth space is realized in this section. We first formulate our problem by adapting it to the background of cGAN, then the detailed architecture design is described and discussed.

\subsection{Problem Formulation}
First proposed in \cite{goodfellow2014generative}, GAN has achieved impressive results in a wide variety of generative tasks. The core idea of GAN is to train two neural networks, which respectively represent the generator \textit{G} and the discriminator \textit{D}, to proceed a game-theoretic tussle between one another. Given the samples $x$ from the real data distribution $p_{data}(x)$ and random noise $z$ sampled from a noise distribution $p_{z}(z)$, the discriminator aims to distinguish between real samples $x$ and fake samples which are mapped from $z$ by the generator, while the generator is tasked with maximally confusing the discriminator. The objective can thus be written as:
\begin{equation}\label{GAN_loss}
\mathcal{L}_{GAN}(G,D) = \mathbb{E}_{x{\sim}p_{data}(x)}[\log D(x)] + \mathbb{E}_{z{\sim}p_{z}(z)}[\log (1-D(G(z)))]
\end{equation}
where $\mathbb{E}$ denotes the empirical estimate of expected value of the probability. To optimize this loss function, we aim to minimize its value for \textit{G} and maximize it for \textit{D} in an adversarial way, i.e. $\min_G \max_D \mathcal{L}_{GAN}(G,D)$. 

The advantage of GAN is that realistic images can be generated from noise vectors with random distribution, which is crucially important for unsupervised learning. However, note that in our face recovery scenario, training data contains image pairs $\{x,y\}$ where $x$ and $y$ refer to the depth and color faces respectively with a one-to-one correspondence between them. The fact that $y$ can be involved in the model as a prior for generative task leads us to the conditional variant of GAN, namely cGAN \cite{mirza2014conditional}. Specifically, we condition the observations $y$ on both the discriminator and the generator, the objective of cGAN extends \eqref{GAN_loss} to:
\begin{equation}\label{cGAN_loss}
\mathcal{L}_{cGAN}(G,D) = \mathbb{E}_{x,y{\sim}p_{data}(x,y)}[\log D(x,y)] + \mathbb{E}_{z{\sim}p_{z}(z), y{\sim}p_{data}(y)}[\log (1-D(G(z|y),y))]
\end{equation}

Moreover, to ensure the pixel-wise similarity between image generation outputs $G(z|y)$ and ground truth $x$, we subsequently impose a reconstruction constraint on the generator in the form of L1 distance between them:
\begin{equation}\label{L1_loss}
\mathcal{L}_{L1}(G) = \mathbb{E}_{x,y{\sim}p_{data}(x,y),z{\sim}p_{z}(z)}[\Vert x - G(z|y) \Vert_1]
\end{equation}

The comprehensive objective is formulated with a minmax value function on the above two losses where the scalar $\eta$ is used for balancing them:
\begin{equation}\label{final_loss}
\min_G \max_D [\mathcal{L}_{cGAN}(G,D) + \eta\mathcal{L}_{L1}(G)]
\end{equation}

Note that the cGAN itself can hardly generate specified images and using only $\mathcal{L}_{L1}(G)$ causes blurring, this joint loss successfully leverages the complementary strengths of them. 

\begin{figure*}
\begin{minipage}[c][7.2cm][t]{.3\textwidth}
  \vspace*{0.6cm}
  \hspace*{0.1cm}
  \centering
  \includegraphics[width=\textwidth]{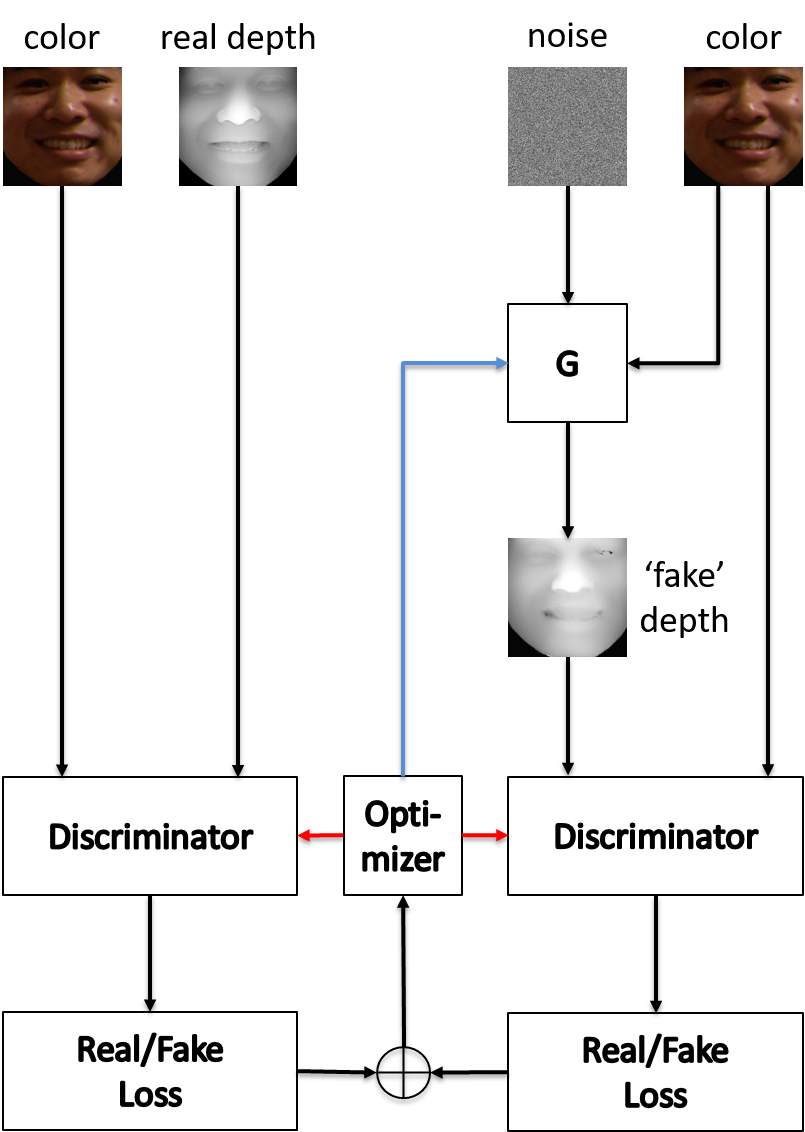}
  \subcaption{Workflow of cGAN}
  \label{fig:test1}
\end{minipage}%
\begin{minipage}[c][7.2cm][t]{.7\textwidth}
  \centering
  \includegraphics[width=.85\textwidth]{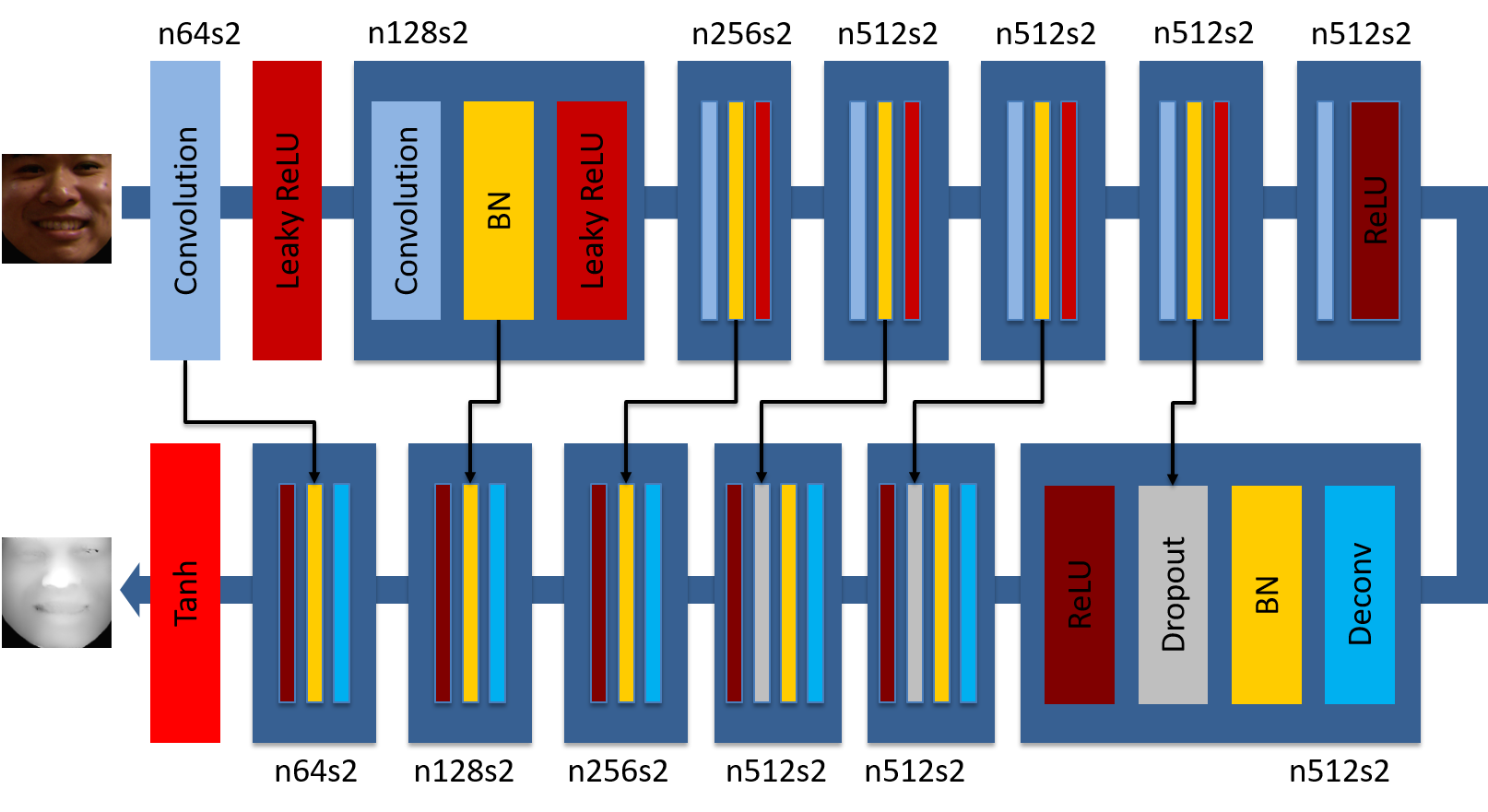}
  \subcaption{Generator}
  \label{fig:test2}\par
  \includegraphics[width=.85\textwidth]{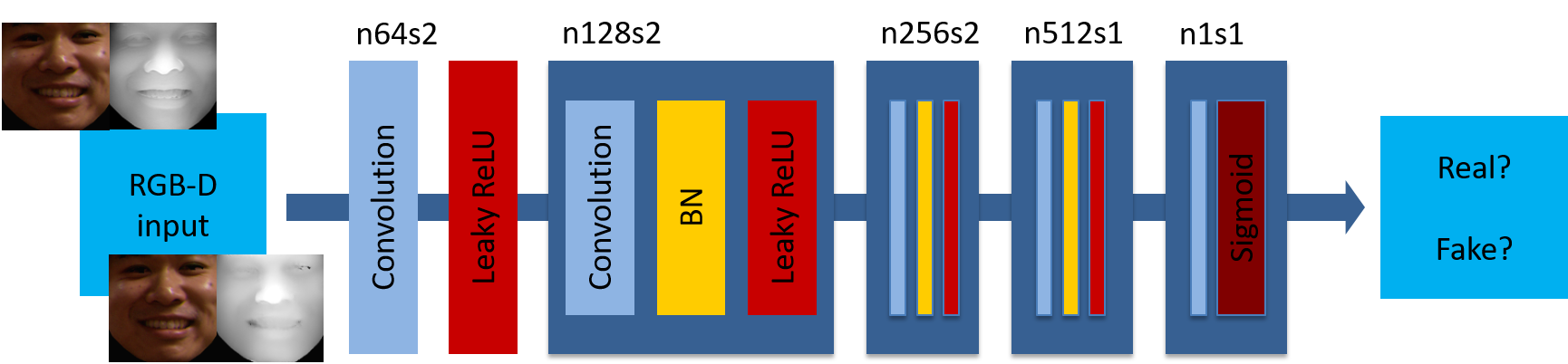}
  \subcaption{Discriminator}
  \label{fig:test3}
\end{minipage}
\caption{The mechanism and architecture of cGAN. In Fig. \ref{fig:test2}, the noise variable $z$ presents itself under the the form of dropout layers, while the black arrows portray the skip connections. All convolution and deconvolution layers are with filter size 4$\times$4 and 1-padding, $n$ and $s$ represent the number of output channels and stride value, respectively. (Best view in color)}
\label{fig:cGAN}
\end{figure*}

\subsection{CGAN Architecture}
We adapt our cGAN architecture from that in \cite{isola2016image} which achieved particularly impressive results in image-to-image translation task. A detailed description of this model is illustrated in Fig. \ref{fig:cGAN} and some key features are discussed below.

\textbf{Generator:} As a standard generative model, the architectures of auto-encoder (AE) \cite{hinton2006reducing} and its variants \cite{vincent2010stacked, rifai2011contractive, kingma2013auto} are widely adopted as $G$ for past cGANs. However, the drawback of conventional AEs is obvious: due to their dimensionality reduction capacity, a large portion of low-level information, such as precise localization, is compressed when an image passes through layers in the encoder. To cope with this lossy compression problem, we follow the idea of U-Net \cite{ronneberger2015u} by adding skip connections which forwards directly the features from encoder layers to decoder layers that are on the same 'level', as shown in Fig. \ref{fig:test2}.

\textbf{Discriminator:} Consistent with Isola et al. \cite{isola2016image}, we adopt \textit{Patch}GAN for the discriminator. Within this pattern, no fully connected layers are implemented and $D$ outputs a 2D image where each pixel represents the prediction result with respect to the corresponding patch on original image. All pixels are then averaged to decide whether the input image is 'real' or 'fake'. Compared with pixel-level prediction, \textit{Patch}GAN efficiently concentrates on local patterns while the global low-frequency correctness is enforced by L1 loss in \eqref{L1_loss}.

\textbf{Optimization:} The optimization for cGAN is performed by following the standard method \cite{goodfellow2014generative}: the mini-batch SGD and the Adam solver are applied to optimize $G$ and $D$ alternately (as depicted by arrows with different colors in Fig. \ref{fig:test1}).

\section{Heterogeneous Face Recognition}
The reconstruction of depth faces from color images enables us to maximally leverage shape information in both gallery and probe, which means we can individually learn a CNN model to extract discriminative features for depth images and transform the initial cross-modal problem into a multi-modal one. However, the heterogeneous matching remains another challenge in our work, below we demonstrate how this problem is formulated and tackled.

\begin{figure}
\centering
\includegraphics[width=0.9\linewidth]{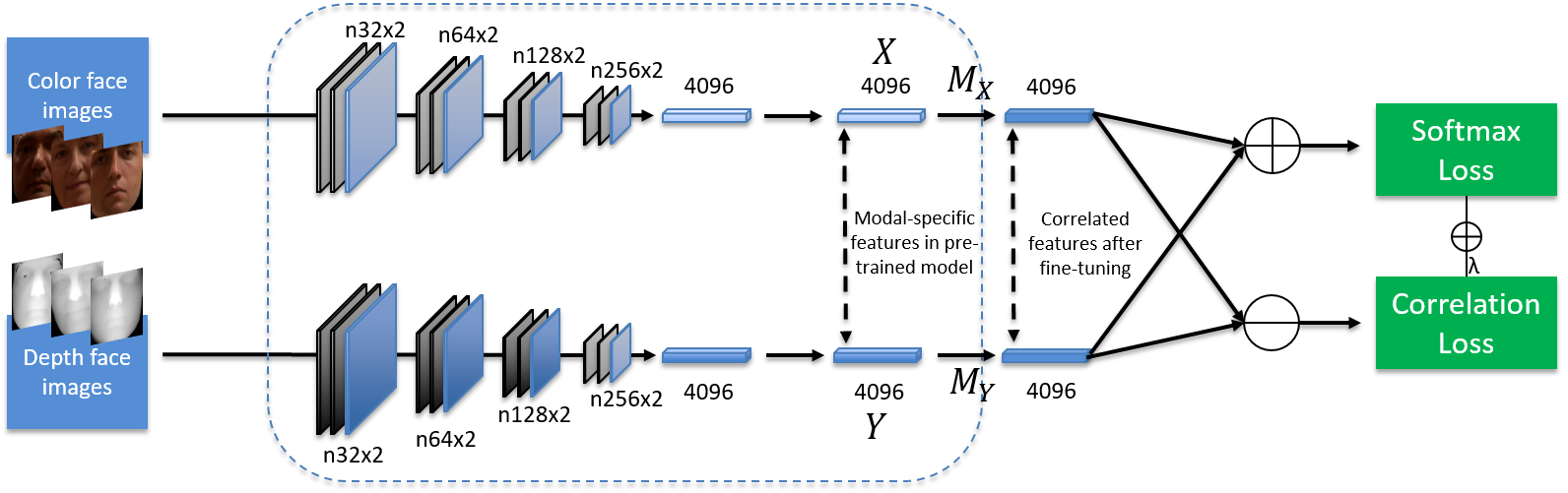}
\caption{Training procedure of the cross-modal CNN model. Models in the dashed box are pre-trained using 2D and 2.5D face images individually. }
\label{fig:hfr}
\end{figure}

\textbf{Unimodal learning.} The last few years witnessed a surge of interest and success in FR with deep learning \cite{taigman2014deepface, sun2015deepid3, parkhi2015deep}. Following the basic idea of stacking convolution-convolution-pooling (C-C-P) layers in \cite{lee2016accurate}, we train from scratch two CNNs for color and grayscale images on CASIA-WebFace \cite{yi2014learning} and further fine-tune the grayscale based model with our own depth images. These two models serve two purposes: to extract 2D and 2.5D features individually, and to offer pre-trained models for the ensuing cross-modal learning.

\textbf{Cross-modal learning.} Once a pair of unimodal models for both views are trained, the modal-specific representations, $\{X,Y\}$, can be obtained after the last fully connected layers. Note that each input for the two-stream cross-modal CNN is a 2D+2.5D image pair with identity correspondence, it is reasonable to have an intuition that $X$ and $Y$ share common patterns which help to classify them as the same class. This connection essentially reflects the nature of cross-modal recognition, and was investigated in \cite{wang2016correlated, huang2012oriented, wang20142d}. 

In order to explore this shared and discriminative feature, a joint supervision is required to enforce both correlation and distinctiveness simultaneously. For this purpose, we apply two linear mappings following $X$ and $Y$, denoted by $M_X$ and $M_Y$. First, to ensure the correlation between new features, they are enforced to be as close as possible, which is constrained by minimizing their distance in feature space:
\begin{equation}\label{frob}
\mathcal{L}_{corr} = \sum_{i=1}^{n}\Vert M_X X_i - M_Y Y_i\Vert_F^2
\end{equation}
where $n$ denotes the size of mini-batch and $\Vert\cdot\Vert_F$ represents the Frobenius norm.

If we only use the above loss supervision signal, the model will simply learn zero mappings for $M_X$ and $M_Y$ because the correlation loss will stably be 0 in this case. To avoid this tricky situation, we average the two outputs to obtain a new feature on which the classification loss is computed. The ultimate objective function is formulated as follows:
\begin{align*}\label{eqhfr}
\mathcal{L}_{hfr} &= \mathcal{L}_{softmax} + \lambda\mathcal{L}_{Corr}\\
&= -\sum_{i=1}^{n}\log\frac{e^{W_{c_i}^T(M_X X_i + M_Y Y_i)/2+b_{c_i}}}{\sum_{j=1}^m e^{W_{j}^T(M_X X_i + M_Y Y_i)/2+b_{j}}} + \lambda\sum_{i=1}^{n}\Vert M_X X_i - M_Y Y_i\Vert_F^2
\end{align*}
where $c_i$ represents the ground truth class label of $i$th image pair, the scalar $\lambda$ denotes the weight for correlation loss.

\textbf{Fusion.} To highlight the effectiveness of the proposed method, we adopt the cosine similarity of 4096-d hidden layer features as matching scores. As for the score fusion stage, all scores are normalized to [0,1] and fused by a simple sum rule.

\section{Experimental Results}
To intuitively demonstrate the effectiveness of the proposed method, we conduct extensive experiments for 2D/2.5D HFR on the benchmark 2D/2.5D face database. Besides the reconstructed 2.5D depth image, our method also outperforms state-of-the-art performance using only 2.5D images instead of holistic 3D face models.

\subsection{Dataset Collection}
Collecting 2D/2.5D image pairs presents itself as a primary challenge when considering deep CNN as a learning pipeline. Unlike the tremendous boost in dataset scale of 2D face images, massive 3D face data acquisition still remains a bottleneck for the development and practical application of 3D based FR techniques, from which our work is partly motivated. 

\textbf{Databases:} As listed in Table \ref{database}, three large scale and publicly available 3D face databases are gathered as training set and the performance is evaluated on another dataset, which implies that there was no overlap between training and test set and the generalization capacity of the proposed method is evaluated as well. Note that the attribute values only concern the data used in our experiments, for example, scans with large pose variations in CASIA-3D are not included here. 

\begin{table}
\begin{center}
\begin{tabular}{ccccc}
\hline
\multirow{2}{*}{Databases} & \multicolumn{3}{c}{Training Set} & Test set \\ \cline{2-4}
& BU3D \cite{yin20063d} 
& Bosphorus \cite{savran2008bosphorus} 
& CASIA-3D \cite{CASIA-3D}
& FRGC Ver2.0 \cite{phillips2005overview}
\\
\hline
\# Persons  & 100  & 105  & 123  & 466 \\
\# Images   & 2500 & 2896 & 1845 & 4003 \\ 
Conditions  & E    & E    & EI   & EI \\          
\hline
\end{tabular}
\end{center}
\caption{Database overview. E and I are short for expressions and illuminations, respectively.}
\label{database}
\end{table}





\textbf{Preprocessing:} To generate 2.5D range image from original 3D shape, we either proceed a direct projection if the point cloud is pre-arranged in grids (Bosphorus/FRGC) or adopt a simple Z-buffer algorithm (BU3D/CASIA3D). Furthermore, to ensure that all faces are of the similar scale, we resize and crop the original image pairs to $128\times128$ while fixing their inter-ocular distance to a certain value. Especially, to deal with the missing holes and unwanted body parts (shoulder for example) in raw data of FRGC, we first locate the face based on 68 automatically detected landmarks \cite{asthana2014incremental}, and then apply a linear interpolation to approximate the default value of each hole pixel by averaging its non-zero neighboring points. 



\subsection{Implementation details} 

All images are normalized before being fed to the network by subtracting from each channel its mean value over all training data. With regards to the choice of hyperparameters, we adopt the following setting: in cGAN, the learning rate $\mu_{cGAN}$ is set to 0.0001 and the weight for L1 norm $\eta$ is 500; in cross-modal CNN model, the learning rate for training from scratch $\mu$ begins with 1 and is divided by 5 every 10 epochs while the learning rate during fine-tuning $\mu_{ft}$ is 0.001; for both models, the momentum $m$ is initially set as 0.5 until it is increased to 0.9 at the 10th epoch; the weight for correlation loss $\lambda$ is set to 0.6.

\subsection{Reconstruction Results}

\begin{figure}
\centering
\begin{subfigure}{.75\textwidth}
\centering
\includegraphics[width=0.9\textwidth]{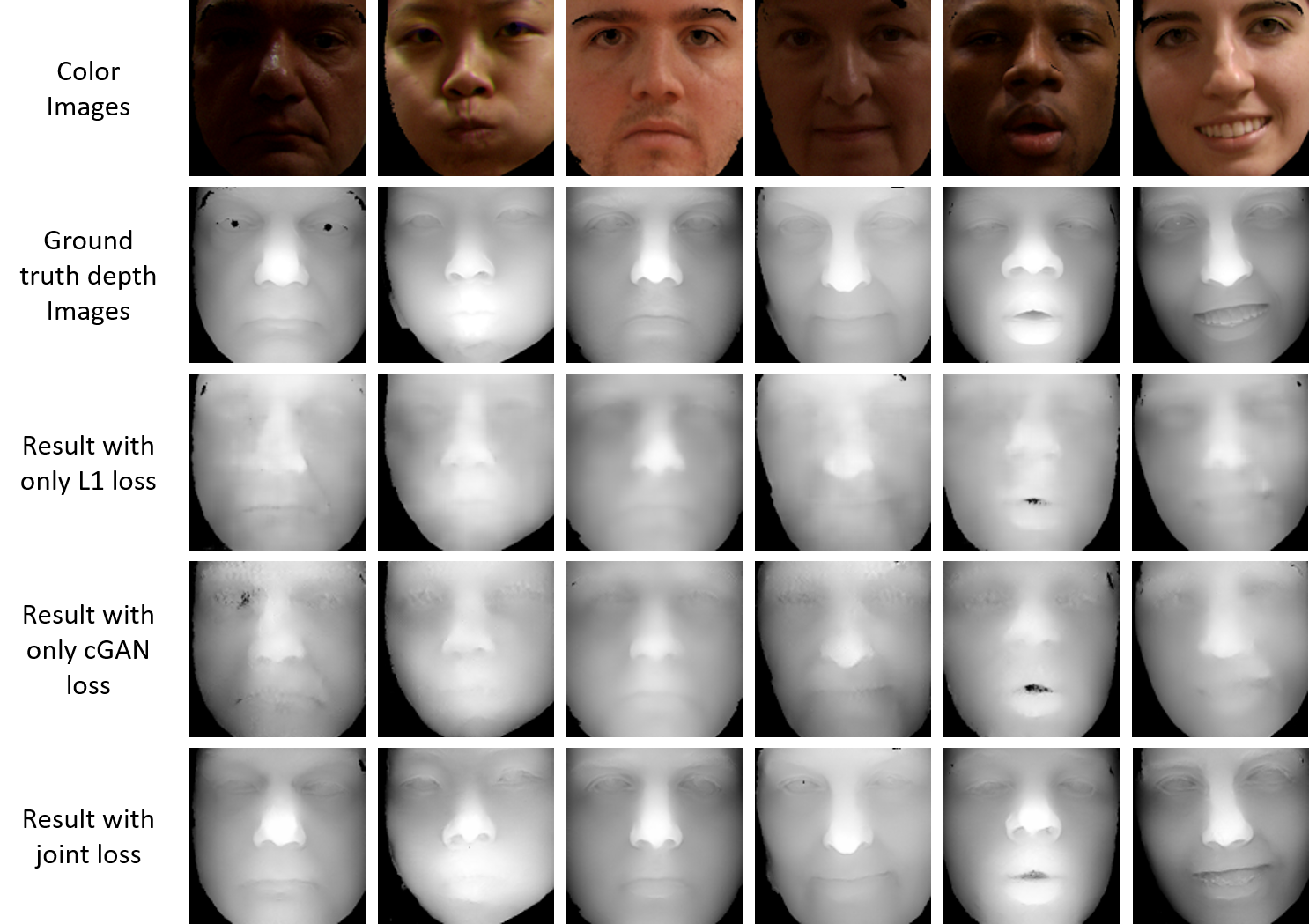}
\caption{}
\label{recovery_good}
\end{subfigure}%
\begin{subfigure}{.25\textwidth}
\centering
\includegraphics[width=0.75\textwidth]{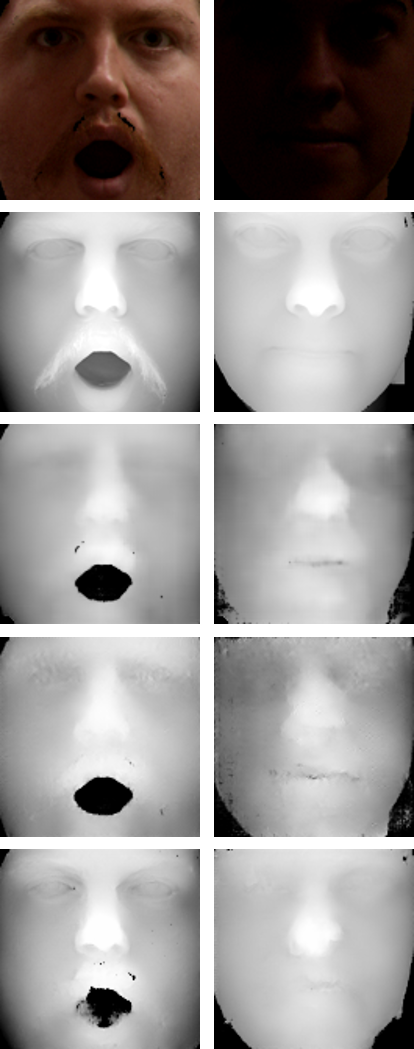}
\caption{}
\label{recovery_bad}
\end{subfigure}
\caption{Qualitative reconstruction results of FRGC samples with varying illuminations and expressions. Fig. \ref{recovery_good}: correctly recovered samples. Fig . \ref{recovery_bad}: wrongly recovered samples.}
\label{recovery}
\end{figure}

The reconstruction results obtained for color images in FRGC are illustrated in Fig. \ref{recovery}. Samples from different subjects across expression and illumination variations are shown from left to right. They thereby give hints on the generalization ability of the proposed method. For each sample we first portray the original color image with its ground truth depth image, followed by the reconstructed results whereby we demonstrate the effectiveness and necessity of each constraint in the joint objective. In addition, some samples with low reconstruction quality are depicted in Fig. \ref{recovery_bad} as well.

Being consistently similar with the ground truth, the reconstruction results with joint loss in Fig. \ref{recovery_good} intuitively demonstrate the strength of cGAN. The recovered depth faces hold their accuracy and realistic property irrespective of lighting and expression variations in the original RGB images. Furthermore, when we take an observation of the two reconstruction results in 3rd and 4th rows, the comparison implies that: 1) using only L1 loss will lead to blurry results because the model tends to average all plausible values, especially for regions containing high-level information like edges; 2) using only cGAN loss can achieve slightly sharper results, but suffers from noise. These results provide an evidence that the implementation of joint loss is beneficial and important for obtaining a 'true' and accurate output. Meanwhile, our model encounters some problems while dealing with extreme cases, such as thick beard, wide opened mouth and extremely dark shadows as displayed in Fig. \ref{recovery_bad}. The errors are principally due to few training samples with these cases.

\subsection{2D-3D Asymmetric FR}

We conduct the quantitative experiments on FRGC which has held the field as one of the most commonly used benchmark dataset over the last decade. In contrast with unimodal FR experiments, very few attempts have been made on 2D/3D asymmetric FR. For convenience of comparison, three recent and representative protocols reported respectively in \cite{jin2014cross}, \cite{huang2012oriented} and \cite{wang20142d} are followed. These protocols mainly differ in gallery and probe setting, including splitting and modality setup. For example, the gallery set in \cite{wang20142d} solely contains depth images, when compared with their work, our experiment will subsequently exclude 2D based matching to respect this protocol.


The comparison results are shown in Table \ref{recognition}, through which we could gain the observation that the proposed cross-modal CNN outperforms state-of-the-art performance while fusing 2.5D matching into HFR with reconstructed depth image further helps improve the performance effectively. Moreover, the proposed method is advantageous in its 3D-free reconstruction capacity and efficiency. To the best of our knowledge, this is the first time to investigate a 2.5D face recovery approach which is free of any 3D prototype models. Despite nearly 20 hours for the whole training and fine-tuning procedure, it takes only 1.6 $ms$ to complete an online forward pass per image on a single NVIDIA GeForce GTX TITAN X GPU and is therefore capable of satisfying the real-time processing requirement. 

\begin{table}
\begin{center}
\begin{tabular}{cccccc}
\hline
\multirow{2}{*}{Protocol} & \multirow{2}{*}{Methods} & \multicolumn{4}{c}{Rank-1 Recognition Accuracy} \\ \cline{3-6}
& & 2D & 2.5D & 2D/2.5D & Fusion \\
\hline
\multirow{2}{*}{Jin et al. \cite{jin2014cross}} & MSDA+ELM \cite{jin2014cross}                                        & -      & - & 0.9680 & - \\
                    & Ours & -      & 0.9573 & 0.9603      & \textbf{0.9698} \\      
\multirow{2}{*}{Wang et al. \cite{wang20142d}} & GRBM+rKCCA \cite{wang20142d}                                            & -      & - & 0.9600 & - \\
                    & Ours & -      & 0.9529 & 0.9714      & \textbf{0.9745} \\ 
\multirow{2}{*}{Huang et al. \cite{huang2012oriented}} & OGM \cite{huang2012oriented}                            & 0.9390 & - & 0.9404 & 0.9537 \\
                    & Ours & 0.9755      & 0.9609 & 0.9688      & \textbf{0.9792} \\ 
\hline
\end{tabular}
\end{center}
\caption{Comparison of recognition accuracy on FRGC under different protocols.}
\label{recognition}
\end{table}

\begin{table}
\begin{center}
\begin{tabular}{cccccccc}
\hline
$\lambda$ & 0 & 0.2 & 0.4 & 0.6 & 0.8 & 1 & 1.2 \\
\hline
Accuracy & 0.9245 & 0.9481 & 0.9600 & 0.9688 & 0.9577 & 0.8851 & 0.7333\\
\hline
\end{tabular}
\end{center}
\caption{2D/2.5D HFR accuracy with varying $\lambda$ under protocol of \cite{huang2012oriented}.}
\label{hyper}
\end{table}

\textbf{Effect of hyperparameter $\lambda$.} An extended analysis is made to explore the role of softmax loss and correlation loss. We take the protocol in \cite{huang2012oriented} as a standard and vary the weight for correlation loss $\lambda$ each time. As shown in Table \ref{hyper}, the performance will remain largely stable across a range of $\lambda_c$ between 0.4 and 0.8. When we set $\lambda=0$ instead of 0.6, which means correlation loss is not involved while training, the network can still learn valuable features with a recognition rate decrease of 4.43$\%$. However, along with the increase of $\lambda$, the performance drops drastically, which implies that a too strong constraint on correlation loss could backfire by causing a negative impact on softmax loss.

\section{Conclusion}
In this paper, we have presented a novel framework for 2D/2.5D heterogeneous face recognition together with depth face reconstruction. This approach combines the generative capacity of conditional GAN and the discriminative feature extraction of deep CNN for cross-modality learning. The extensive experiments have convincingly evidenced that the proposed method successfully reconstructs realistic 2.5D from single 2D while being adaptive and sufficient for HFR. This architecture could hopefully be generalized to other heterogeneous FR tasks, such as visible light vs. near-infrared and 2.5D vs. forensic sketch, which provides an interesting and promising prospect.

\section{Acknowledgement}
This work was supported in part by the French Research Agency, l'Agence Nationale de Recherche (ANR), through the Jemime project (N$^\circ$ contract ANR-13-CORD-0004-02), the Biofence project (N$^\circ$ ANR-13-INSE-0004-02) and the PUF 4D Vision project funded by the Partner University Foundation.

\bibliography{egbib}
\end{document}